\documentclass[10pt,letterpaper]{article}

\usepackage{authblk}
\usepackage{times}
\usepackage{epsfig}
\usepackage{graphicx}
\usepackage[scale=0.8]{geometry}
\usepackage{amsmath}
\usepackage{amssymb}
\usepackage{mathrsfs}
\usepackage{float}
\usepackage{multirow}

\usepackage[breaklinks=true,bookmarks=false]{hyperref}

\begin{document}

\title{Randomized Forward Mode Gradient
for Spiking Neural Networks in Scientific Machine Learning}

\author[1]{Ruyin Wan\thanks{ruyin\_wan@brown.edu}}
\author[2]{Qian Zhang\thanks{qian\_zhang1@brown.edu}}
\author[1, 2]{George Em Karniadakis\thanks{george\_karniadakis@brown.edu}}

\affil[1]{School of Engineering, Brown University, Providence, RI, 02912}
\affil[2]{Division of Applied Mathematics, Brown University, Providence, RI, 02912}

\date{}

\maketitle

\begin{abstract}

Spiking neural networks (SNNs) represent a promising approach in machine learning, combining the hierarchical learning capabilities of deep neural networks with the energy efficiency of spike-based computations. Traditional end-to-end training of SNNs is often based on back-propagation, where weight updates are derived from gradients computed through the chain rule. However, this method encounters challenges due to its limited biological plausibility and inefficiencies on neuromorphic hardware. In this study, we introduce an alternative training approach for SNNs. Instead of using back-propagation, we leverage weight perturbation methods within a forward-mode gradient framework. Specifically, we perturb the weight matrix with a small noise term and estimate gradients by observing the changes in the network output. Experimental results on regression tasks, including solving various PDEs, show that our approach achieves competitive accuracy, suggesting its suitability for neuromorphic systems and potential hardware compatibility.
\end{abstract}

\section{Introduction}

Recent advances in machine learning have greatly expanded the capabilities of artificial intelligence (AI) for applications in solving differential equations, function approximation, and various other fields \cite{jin2020sympnets, cai2022pinns, cao2023deep, toscano2023teeth, zhang2022aoslo}. As demand grows for energy-efficient and computationally feasible solutions, spiking neural networks (SNNs) have attracted significant attention within the machine learning community. Unlike traditional artificial neural networks (ANNs), which rely on continuous activation functions, SNNs operate with sparse, binary spiking events, making them more efficient in terms of energy consumption \cite{maass1997networks, massa2020efficient, kim2022rate, bouvier2019spiking, roy2019towards}. Furthermore, SNNs benefit from recent advancements in neuromorphic hardware (e.g., Intel's Loihi 2 chip \cite{davies2018loihi, orchard2021efficient}), which enables SNNs to approximate brain-like computations, hence facilitating lightweight faster models.

Despite these benefits, most deep neural networks still rely on back-propagation for training. This method, however, is considered ``biologically implausible" as it lacks symmetry with the way biological neural systems learn, does not engage massive parallelism, and is often incompatible with neuromorphic hardware. This incongruity underscores the need for novel learning algorithms better suited to SNNs.

Weight perturbation offers a promising alternative, where small perturbations are applied to the synaptic connections during the forward pass, and weight updates are adjusted in response to changes in the loss function. Instead of perturbing weights directly, forward-mode automatic differentiation (AD) can be employed to compute a directional gradient along the perturbation direction \cite{pearlmutter1994fast}. Forward-mode AD has seen a resurgence in deep learning applications \cite{silver2022learning}.

In this work, we leverage weight perturbation to train SNNs, aiming for greater biological plausibility. We explore two different methods for determining surrogate gradients and implementing perturbations. Evaluating these methods on regression tasks that are more changeling \cite{zhang2022sms, kahana2022spiking, zhang2023artificial}, particularly relevant to scientific machine learning (SciML), our results indicate the viability of this approach, positioning it as a step towards realizing SNNs on neuromorphic hardware~\cite{theilman2024loihi} without reliance on back-propagation.

The paper is organized as follows. In section 2, we review related works, and in section 3 we describe the methodology. In section 4 we present our results, and we conclude in section 5 with a summary. 

\section{Related Works}

\begin{figure}[t]
\begin{center}
  \includegraphics[width=0.4\textwidth,keepaspectratio]{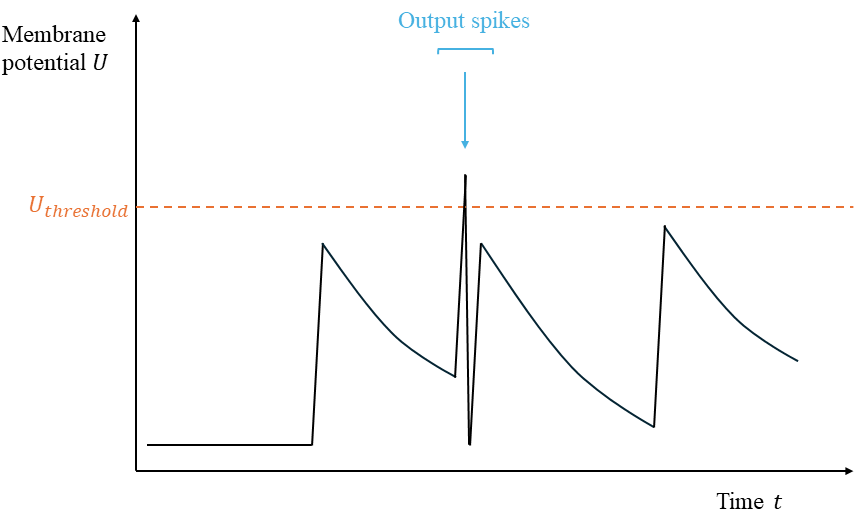}
\end{center}
  \caption{Membrane activity and spike generation.}
\label{fig:membrane}
\end{figure}

\begin{figure}[t]
\begin{center}
  \includegraphics[width=0.6\textwidth,keepaspectratio]{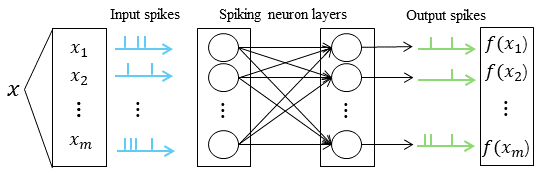}
\end{center}
  \caption{Spiking  multi-layer perceptron  (MLP).}
\label{fig:spike_mlp}
\end{figure}

\textbf{Spiking Neural Networks (SNNs).} SNNs simulate biological neurons by processing spiking data through membrane potentials and synapses, capturing time-dependent binary information. The Leaky Integrate-and-Fire (LIF) model \cite{lapique1907recherches} is a widely adopted framework for emulating neuron membrane dynamics as a Resistor-Capacitor (RC) circuit. Illustrated in Figure ~\ref{fig:membrane}, the LIF neuron accumulates membrane voltage over time, emitting a spike once the voltage surpasses a pre-defined threshold ($U_{threshold}$). After a spike, the membrane potential resets, and the voltage decays over time, mimicking biological leakage. Synaptic behaviors are typically modeled with fully connected layers in SNNs. Figure ~\ref{fig:spike_mlp} shows the general architecture of a spiking multi-layer perceptron (MLP).

The non-differentiable nature of spike-based operations complicates SNN training. Two primary strategies are widely used: indirect and direct training. Indirect training, or ANN-to-SNN conversion, involves training an ANN and then converting it to an SNN. This approach leverages the accuracy of ANNs and has shown success across architectures \cite{han2020rmp, sengupta2019going, li2021free} and applications \cite{kim2020spiking}. However, conversion may fail to exploit SNNs' sparsity advantages and remains computationally expensive. Direct training, by contrast, working more closely with spikes, employs surrogate gradient methods to approximate gradients and usually propagates backward for model updates, which can be biologically implausible and difficult to deploy on neuromorphic chips.

\textbf{Randomized Forward-Mode Gradient.} Forward-mode automatic differentiation (AD) \cite{wengert1964simple} contrasts with reverse-mode AD by computing gradients through directional derivatives, only using a forward evaluation. This method enables efficient gradient computation via Jacobian-Vector Products (JVP). The gradient estimator proposed by \cite{baydin2022gradients} computes gradients by applying a standard normal random vector in the gradient direction. The authors in \cite{shukla2023randomized} generalized this estimator with flexible distribution choices, introducing the ``randomized forward-mode gradient" (RFG) as a more adaptable method.

\textbf{Biological Plausibility of Perturbation Learning.} In biological learning contexts, forward gradient methods have conceptual links to perturbation learning. Neural plasticity rules focus on weight updates derived from neuron input and output activity \cite{hebb1949organization, widrow1960adaptive, oja1982simplified, bienenstock1982theory, abbott2000synaptic}. Spike-based formulations for weight perturbation learning were proposed by \cite{xie1999spike}, offering a biologically-inspired approach to synaptic update rules.

\textbf{DeepONet and SepONet.} 
The deep operator network (DeepONet) \cite{lu2021learning} is a neural network architecture designed to learn mappings between function spaces, effectively approximating complex operators. Unlike traditional neural networks that learns functions, DeepONet approximates operators \(\mathcal{G}: \mathcal{X} \to \mathcal{Y}\) where \(\mathcal{X}\) and \(\mathcal{Y}\) are spaces of functions, such as solutions to differential equations. DeepONet achieves this by learning an operator \(\mathcal{G}\) that maps a given input function \(u(x) \in \mathcal{X}\) to an output function \(v(y) = \mathcal{G}(u)(y)\), where \(x\) and \(y\) are spatial coordinates.

\begin{figure}[H]
    \centering
    \includegraphics[width=0.7\linewidth]{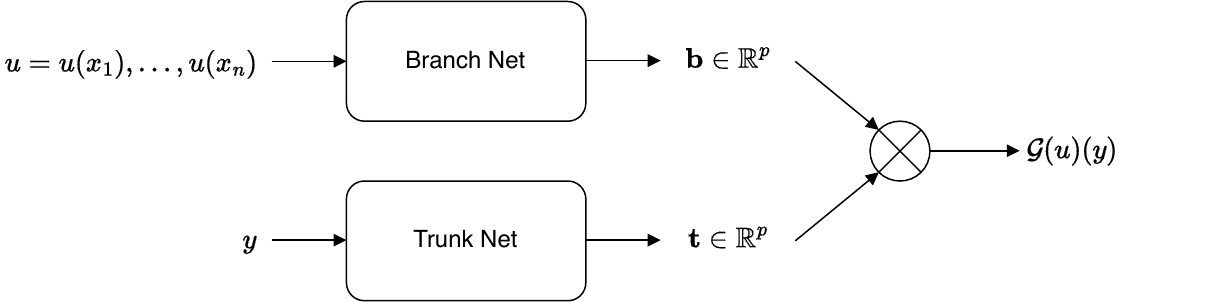}
    \caption{Architecture of DeepONet.}
    \label{fig:deeponet}
\end{figure}

DeepONet decomposes the operator into two networks: a branch network \(\text{Branch}(u)\) that processes the input function \(u(x)\) and a trunk network \(\text{Trunk}(y)\) that processes the target location \(y\). The structure of DeepONet is depicted in Figure \ref{fig:deeponet}. The output of DeepONet is then expressed as:
\begin{equation}
    \mathcal{G}(u)(y) \approx \sum_{i=1}^{p} \text{Branch}_i(u) \cdot \text{Trunk}_i(y),
\end{equation}
where \(m\) is the number of basis functions learned by the network. In practice, the input function \(u(x)\) is discretized to a finite set of points \(\{u(x_j)\}_{j=1}^n\), which are fed into the branch network. The target \(y\) is an arbitrary point in the output domain, where \(\mathcal{G}(u)(y)\) is evaluated by combining the outputs from both the branch and trunk networks. In this paper, we use two spiking MLPs, as illustrated before for branch and trunk network.

To train the DeepONet, a mean squared error loss is commonly used:
\begin{equation}
    \mathcal{L} = \frac{1}{N} \sum_{k=1}^N \left| \mathcal{G}(u^{(k)})(y^{(k)}) - \sum_{i=1}^p \text{Branch}_i(u^{(k)}) \cdot \text{Trunk}_i(y^{(k)}) \right|^2,
\end{equation}
where \(N\) is the number of training samples, and \((u^{(k)}, y^{(k)})\) are sampled pairs of input functions and corresponding output locations. By minimizing this loss, DeepONet learns an accurate approximation to the target operator \(\mathcal{G}\), making it a powerful tool for applications in scientific computing and physics-informed machine learning.

\begin{figure}[H]
    \centering
    \includegraphics[width=0.7\linewidth]{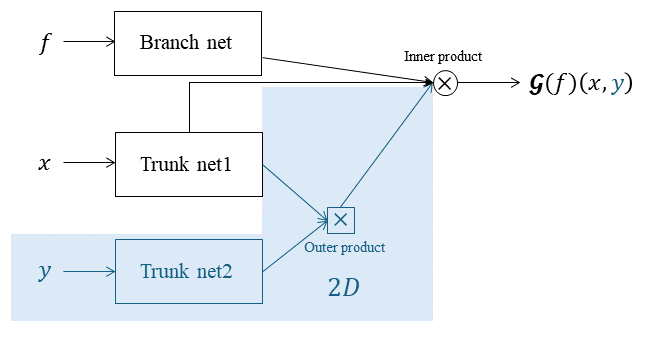}
    \caption{Architecture of SepONet.}
    \label{fig:seponet}
\end{figure}

The Separable Operator Network (SepONet) \cite{yu2024separable}, while still composed of branch and trunk networks, differs from the original DeepONet structure by introducing $d$ independent trunk networks for handling $d$-dimensional problems, rather than using a single trunk network for all dimensions. The structure of SepONet for a two-dimensional problem is illustrated in Figure ~\ref{fig:seponet}. The blue components highlight the addition of an independent trunk network for each dimension. The output of SepONet is formulated as follows:
$$\mathcal{G}(u)(\mathbf{y}) = \mathcal{G}(u)(y_1, y_2, \dots, y_d) \approx \sum_{i=1}^{r} \text{Branch}_i(u) \cdot \prod_{n=1}^d \text{Trunk}_{i, n}(y_n),$$
where $\text{Trunk}_{i, n}: \mathbb{R} \to \mathbb{R}^r$ for $n = 1, 2, \dots, d$ represents the independent trunk network for each dimension. The outer product is taken over each dimension $n = 1, 2, \dots, d$ for the trunk networks, followed by an inner product with the branch network’s prediction, to approximate the target operator $\mathcal{G}$. In this work, we adopt a spiking version of SepONet by implementing spiking MLPs for both the branch network and each trunk network.

\section{Methods}

\textbf{Surrogate gradient for SNNs.} In the $l_{th}$ layer and at time step $t$, the LIF neuron model for SNN is described as follows.  The membrane potential $\mathbf{U}^l$ at time step $t$ is updated by
$$\mathbf{U}^l[t] = \lambda\mathbf{U}^l[t-1] + \mathbf{Z}^l(t),$$
where $\lambda$ is the leak constant, and 
$$\mathbf{Z}^l[t] = \mathbf{W}^l \mathbf{S}^{l-1}[t] + \mathbf{B}^l$$
are the postsynaptic neurons. We employ the Integrate-and-Fire (IF) neuron model without incorporating the leaky constant $\lambda$. In this framework, $\mathbf{S}^{l-1}$ represents the presynaptic spikes in the $l^{th}$ layer, while $\mathbf{W}^l$ denotes the synaptic weights connecting the $(l-1)^{th}$ layer to the $l^{th}$ layer. Additionally, $\mathbf{B}^l$ is the synaptic bias at the $l^{th}$ layer. The spikes $\mathbf{S}$ are generated when the accumulated membrane potential surpasses a defined threshold, $\mathbf{U}_{\text{threshold}}$, at which point information is transmitted across synapses. This process can be expressed as:
$$\mathbf{S}^{l-1}(t) = H(\mathbf{U}^{l-1}[t] - \mathbf{U}_{threshold}),$$
where $H$ is the Heaviside step function given by
$$H(x) = \begin{cases} 
0 & \text{for } x < 0 \\
1 & \text{for } x \geq 0
\end{cases}.$$
After the spike is generated, the membrane potential is reset to
$$\mathbf{U}^{l}[t] = \mathbf{U}^{l}[t] - \mathbf{S}^{l}[t]\mathbf{U}_{threshold}.$$ 

To update the model parameters, we need to compute the gradient of the loss function $\mathcal{L}$ with respect to the parameters, where the loss quantifies the error between the network's output and the labeled data. Given that the recurrent equations defining the network's output involve the non-differentiable Heaviside function, we approximate its gradient with a surrogate. In this work, we employ two approaches to define the surrogate gradient. The first is the standard surrogate gradient
$$f_{sg}(x) = \frac{1}{\sigma\sqrt{2\pi}} e^{-\frac{x^2}{2\sigma^2}} \dot{x}$$
where $\sigma$ is the variance, and the other one is using Stein's lemma \cite{he2023learning} as a weak form of surrogate gradient
$$f_{wsg}(x) = \frac{1}{K} \sum_{k=1}^K \left[ \frac{\delta_k}{\sigma^2} (H(x + \delta_k) - H(x)) \right],$$
where $\delta_k$ are i.i.d. samples from $\mathcal{N}(0, \sigma^2 I)$.


\textbf{Gradient calculation in SNNs.} There are different ways to compute the gradient of the loss. In back-propagation, we compute the gradient with respect to all the parameters $\theta^l$ in $l_{th}$ layer as
$$g_{bp}(\theta^l) = \frac{\partial \mathcal{L}}{\partial \theta^l}$$
$$= \sum_{t} \frac{\partial \mathcal{L}}{\partial \mathbf{S}^{l}[t]} \frac{\partial \mathbf{S}^{l}[t]}{\partial \mathbf{U}^{l}[t]} \frac{\partial \mathbf{U}^{l}[t]}{\partial \theta^l}.$$
Here, 
$$\frac{\partial \mathbf{U}^{l}[t]}{\partial \theta^{l}} = \frac{\partial \mathbf{U}^{l}[t-1]}{\partial \theta^{l}} + \frac{\partial \mathbf{Z}^{l}[t]}{\partial \theta^{l}}$$
and 
$$\frac{\partial \mathbf{Z}^{l}[t]}{\partial \theta^{l}} = \frac{\partial \mathbf{W}^l(\theta^l) \mathbf{S}^{l-1}[t] + \mathbf{B}^l(\theta^l)}{\partial \theta^{l}},$$
so that we can compute recursively.
In randomized forward mode gradient, using perturbations, we have the gradient with respect to $\theta^l$ calculated as
$$g_{rfg, g}(\theta^l) = g_{rfg, g}(\theta)= \sum_{t=1}^{T} \left( \frac{\partial \mathcal{L}}{\partial \theta} \cdot (\delta \theta) \right) (\delta \theta)^l$$
$$= \sum_{t}  \left[ \sum_{n=1}^{N} \left( \frac{\partial \mathcal{L}}{\partial \mathbf{S}^{n}[t]} \frac{\partial \mathbf{S}^{n}[t]}{\partial \mathbf{U}^{n}[t]} \frac{\partial \mathbf{U}^{n}[t]}{\partial \theta^{n}}(\delta \theta)^{n} \right) \right] (\delta \theta)^l,$$
where $(\delta \theta)$ is the random perturbation vector sampled from a probability distribution $\mathbf{P}$ and here we use the Bernoulli distribution and $(\delta \theta)^l$ is the part of the perturbation vector for parameters $\theta^l$ in the $l_{th}$ layer. Forward-mode AD is used to calculate the inner product $\sum_{n=1}^{N} \left( \frac{\partial \mathcal{L}}{\partial \mathbf{S}^{n}[t]} \frac{\partial \mathbf{S}^{n}[t]}{\partial \mathbf{U}^{n}[t]} \frac{\partial \mathbf{U}^{n}[t]}{\partial \theta^{n}}(\delta \theta)^{n} \right)$ where $n$ is the parameter number. In RFG, we could also do a layer-wise perturbation instead of the global perturbation above. That is 
$$g_{rfg,l}(\theta^l) = \sum_{t} \left( \frac{\partial \mathcal{L}}{\partial \theta^l} \cdot (\delta \theta)^l \right) (\delta \theta)^l$$
$$= \sum_{t}  \left[ \sum_{n_l=1}^{N_l} \left( \frac{\partial \mathcal{L}}{\partial \mathbf{S}^{n_l}[t]} \frac{\partial \mathbf{S}^{n_l}[t]}{\partial \mathbf{U}^{n_l}[t]} \frac{\partial \mathbf{U}^{n_l}[t]}{\partial \theta^{n_l}}(\delta \theta)^{n_l} \right) \right] (\delta \theta)^l$$
for layer $l$ and forward-mode AD is also used here to calculate the inner product $\sum_{n_l=1}^{N_l} \left( \frac{\partial \mathcal{L}}{\partial \mathbf{S}^{n_l}[t]} \frac{\partial \mathbf{S}^{n_l}[t]}{\partial \mathbf{U}^{n_l}[t]} \frac{\partial \mathbf{U}^{n_l}[t]}{\partial \theta^{n_l}}(\delta \theta)^{n_l}\right)$, where $n_l$ is the parameter number in the $l_{th}$ layer. This forward gradient computation could be done in parallel for all layers. 
Here, $\frac{\partial \mathbf{S}[t]}{\partial \mathbf{U}[t]}$ is approximated by the surrogate gradient $f(\mathbf{U}^l[t])$. 
After computing the gradient, the parameters $\theta^l$ are updated as
$$\theta^l = \theta^l - \eta g(\theta^l), $$
where $g(\theta^l)$ is the gradient with respect to $\theta^l$ and $\eta$ is the learning rate.


\section{Experiments}
To show the effectiveness of our proposed algorithm, we compare it to different combinations of methods. We explain the methods in detail below.

\textbf{1) Surrogate gradient.} Two different forms of surrogate gradients, the usual surrogate gradient (SG) and the weak form of the surrogate gradient using Stein's formula (WSG) are used. The parameter $\sigma$ is 0.3 for the one-dimensional Poisson equation and 0.5 for other cases.

\textbf{2) Gradient calculation.} We include two different ways of gradient calculation: \textbf{back-propagtion (BP)} and \textbf{randomized forward mode gradient (RFG)}. 

\textbf{3) Perturbation.} \textbf{Global perturbation (G)} and \textbf{layer-wise perturbation (L)} are combined with different gradient calculation methods for comparison.

We denote the combinations of these methods as follows: Surrogate gradient - method of gradient calculation - perturbation. For instance, we use \textbf{WSG-RFG-G} to represent that using Stein's formula as a weak surrogate gradient and perturbing globally. 

\begin{figure}[H]
    \centering
    \begin{tabular}{ccc}
        \includegraphics[width=0.3\textwidth]{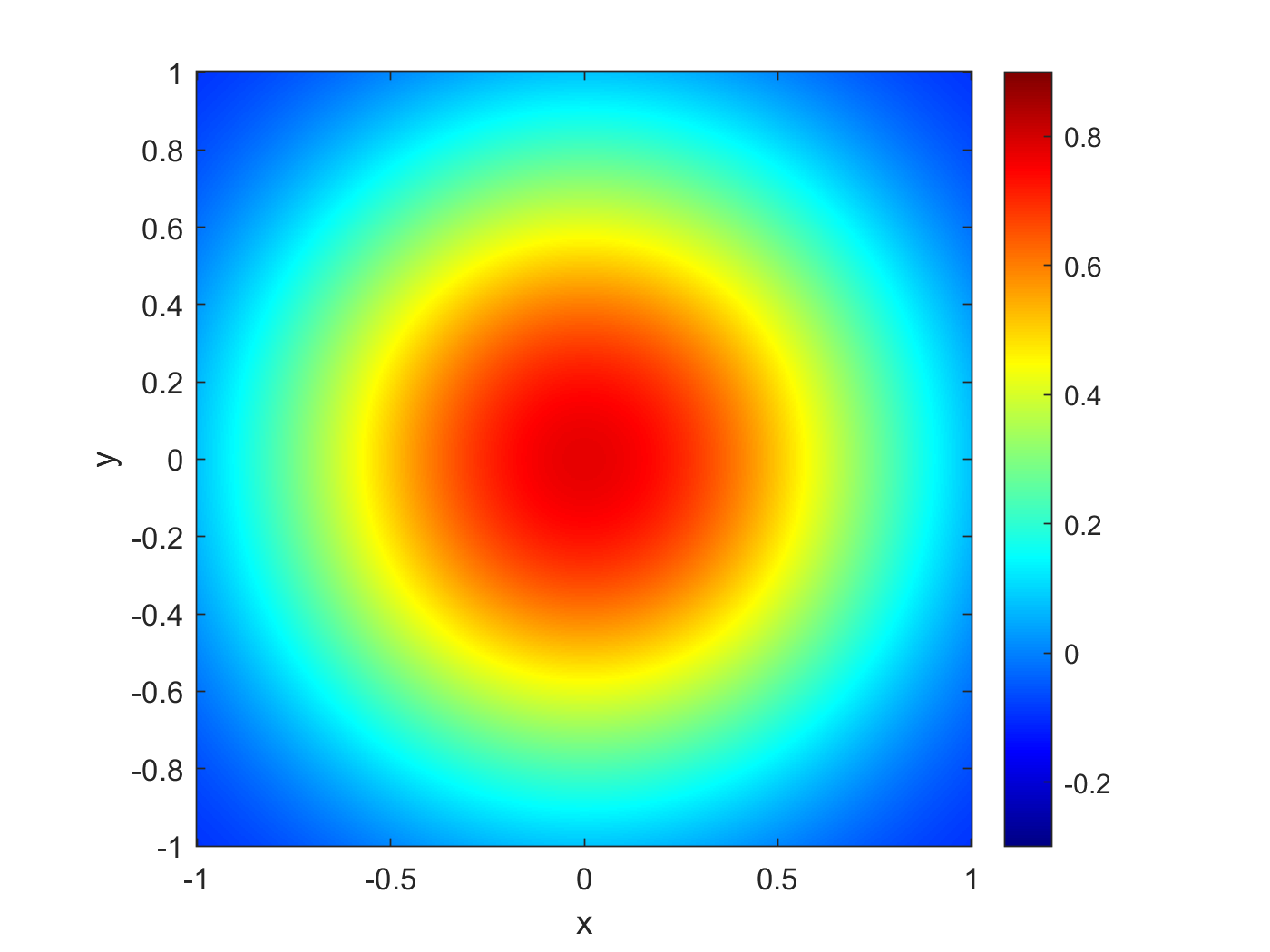} &
        \includegraphics[width=0.3\textwidth]{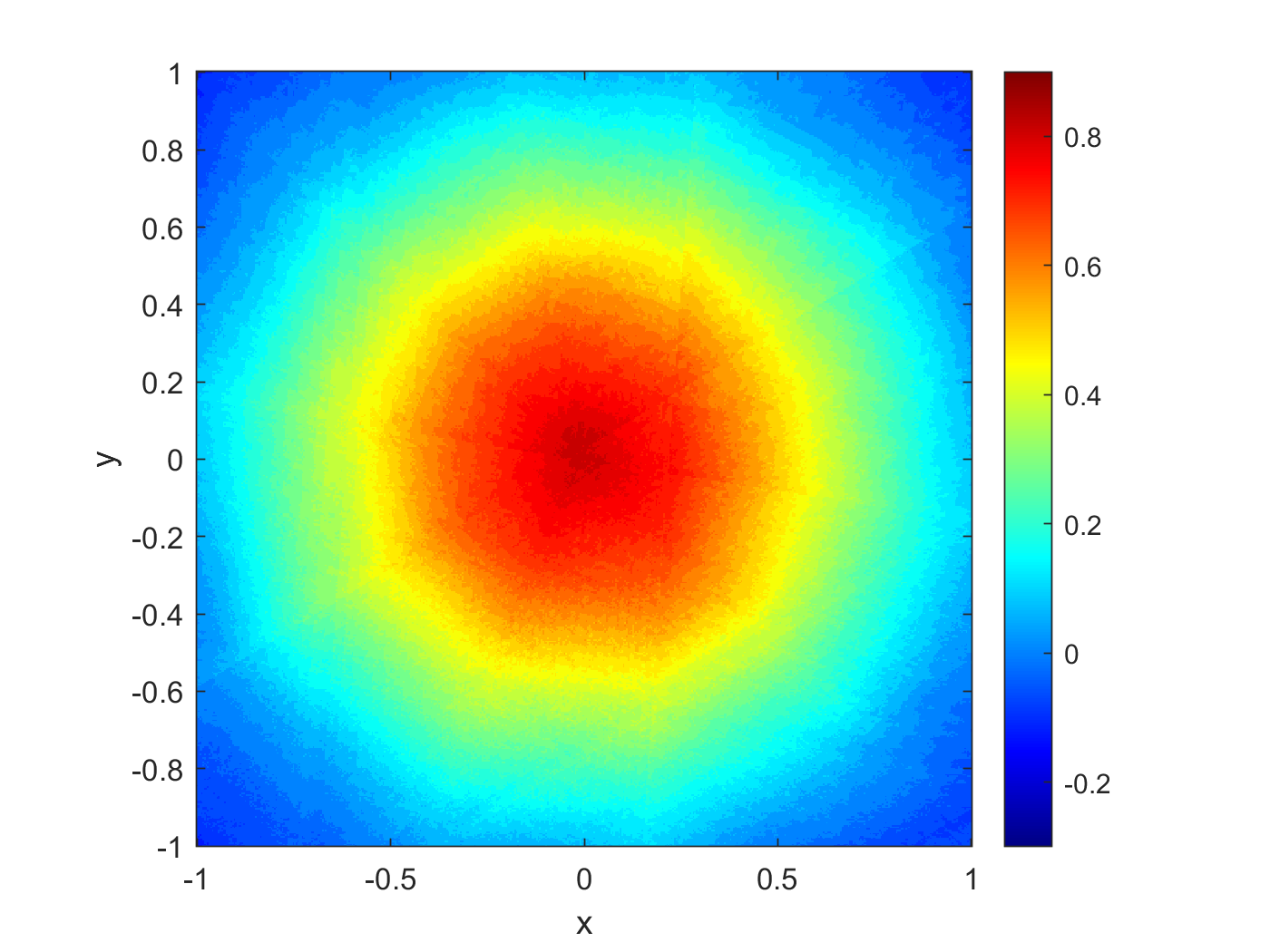} \\
        (a) Ground truth &  (b) Prediction(SG-RFG-G) \\
        \includegraphics[width=0.3\textwidth]{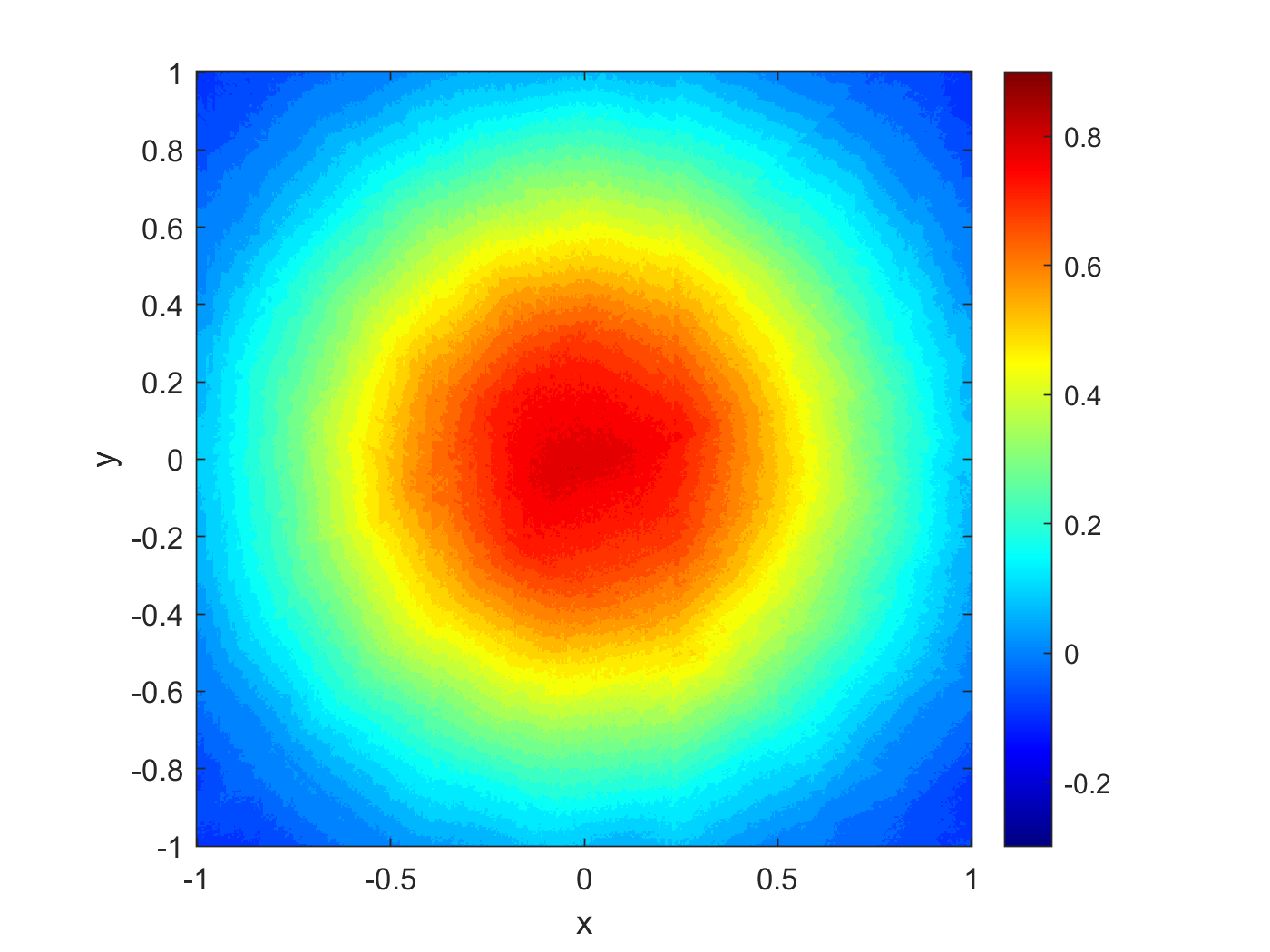} &
        \includegraphics[width=0.3\textwidth]{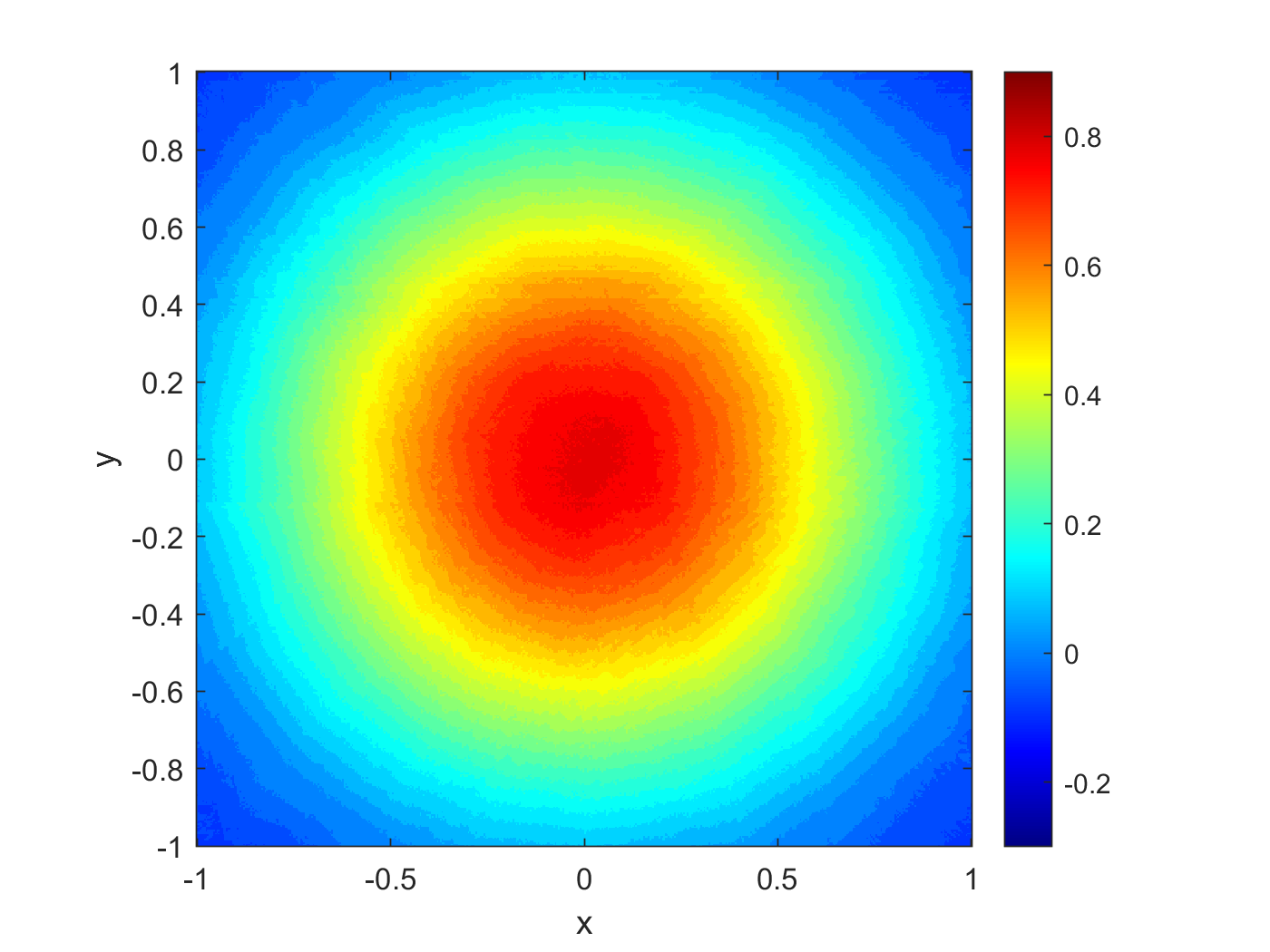}\\ 
        (c) Prediction(SG-RFG-L) & (d) Prediction(SG-BP) \\
    \end{tabular}
    \caption{Function regression on 2D Mexican hat wavelet.}
    \label{fig:2d}
\end{figure}

\textbf{Function regression.} 
By using 2 hidden layers with 16 neurons in each layer and 32 simulation time steps, we approximate the two-dimensional Mexican hat wavelet given by 
$$\psi(x, y) = \frac{1}{\pi \sigma^4} (1-\frac{1}{2}(\frac{x^2+y^2}{\sigma^2})) e^{-\frac{1}{2}(\frac{x^2+y^2}{\sigma^2})}$$
with $\sigma = 0.8$. We generate 10,000 training pairs, $\{x_n, y_n\}$, uniformly distributed over the domain $[-1, 1] \times [-1, 1]$. The results, shown in Figure \ref{fig:2d}, compare the ground truth with the function predictions obtained using randomized forward gradients with global perturbation, layer-wise perturbation, and back-propagation. The corresponding relative $l_2$ errors are 0.0426, 0.0342, and 0.0298, respectively.

\begin{figure}[H]
    \centering
    \begin{tabular}{ccc}
        \includegraphics[width=0.3\textwidth]{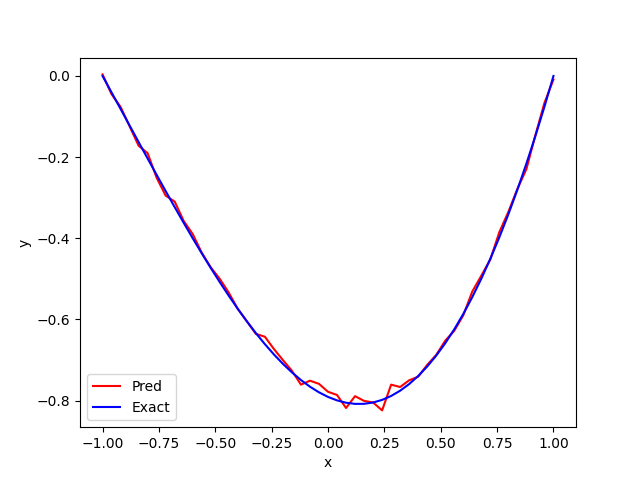} &
        \includegraphics[width=0.3\textwidth]{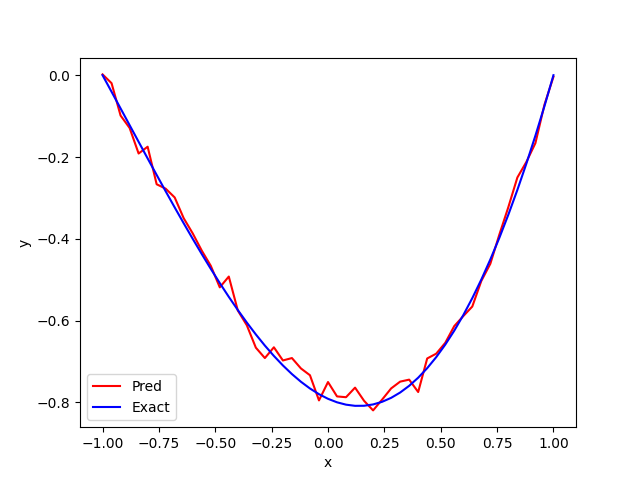} &
        \includegraphics[width=0.3\textwidth]{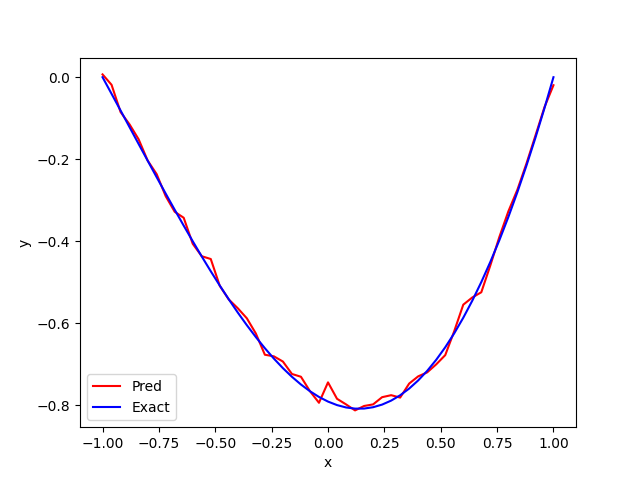} \\
        (a) SG-BP &  (b) SG-RFG-G & (c) SG-RFG-L\\
         \includegraphics[width=0.3\textwidth]{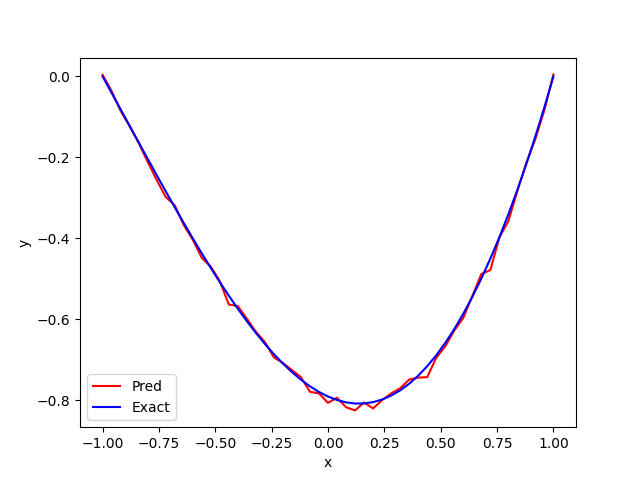} &
        \includegraphics[width=0.3\textwidth]{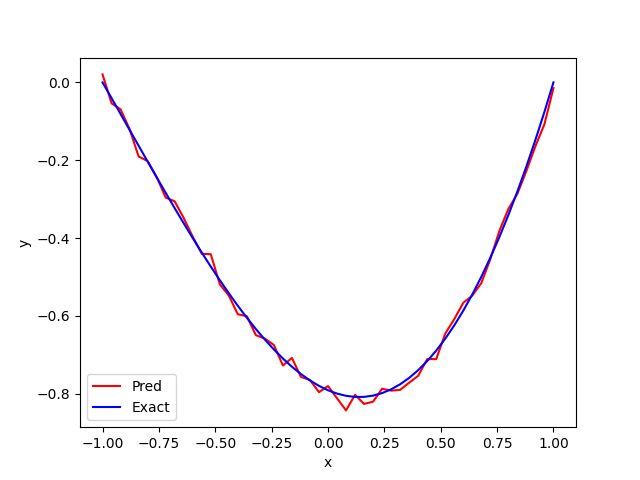} &
        \includegraphics[width=0.3\textwidth]{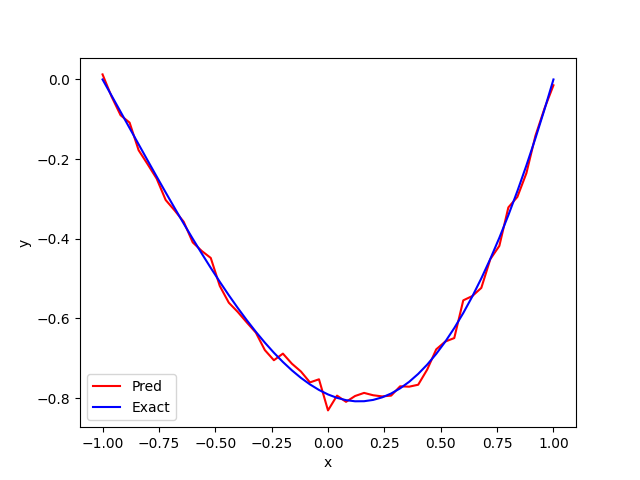} \\
        (d) WSG-BP &  (e) WFG-RFG-G & (f) WSG-RFG-L
    \end{tabular}
    \caption{Solution of 1D Poisson equation.}
    \label{fig:poisson}
\end{figure}

\textbf{Operator learning.}
We solve a 1D Poisson equation 
$$-\frac{d^2 u}{dx^2} = g(x), \quad x \in (-1,1)$$ by the spiking DeepONet. Function $g$ is sampled from a Gaussian Random Field (GRF); 800 samples are used for training and 800 samples for testing. The DeepONet has only 1 hidden layer of width 32 for both branch net and trunk net and the number of simulation time steps is 32. The results are given in Table \ref{1dpoisson}. The plots for one sample prediction using different methods are in Figure \ref{fig:poisson}.

\textbf{Remark.} Due to the difficulties in processing spike information through the neural network, we tried another method such that the loss function includes not only the global loss from the final output, but also composed of the local loss made by the predictions in intermediate layers. The loss function is formulated as 
$$\mathcal{L} = \mathcal{L}_{global}+ \frac{1}{N} \sum_{n=1}^N \mathcal{L}_{n},$$
where $\mathcal{L}_{n}$ is the local loss calculated between a projection of the $n_{th}$ layer output and the target output. By doing this, we expected to force each layer to have some information from the target output. The results are shown in Table \ref{1dpoisson} ending with LL. However, the results showed that this approach is not effective either in back-propagation or RFG for this case, and hence further investigation is required.

\begin{table}[H]
\centering
\caption{Relative $L_2$ error for 1D Poisson equation}\label{1dpoisson}
\begin{tabular}{ccc}
\hline
\textbf{Gradient Calculation} & \textbf{Method Combination} & \textbf{Test Relative $L_2$ Error} \\ \hline
\multirow{4}{*}{Back-propagation} & SG-BP & 0.0246 \\
 & WSG-BP & 0.0241 \\ 
 & SG-BP-LL & 0.0264 \\ 
 & WSG-BP-LL & 0.0244 \\ \hline
\multirow{6}{*}{Randomized Forward Mode Gradient} & SG-RFG-G & 0.0422 \\
 & WSG-RFG-G & 0.0412 \\
 & SG-RFG-L & 0.0347 \\
 & WSG-RFG-L & 0.0359 \\
& SG-RFG-LL & 0.0431 \\
& WSG-RFG-LL & 0.0446 \\\hline
\end{tabular}
\end{table}

\begin{figure}[H]
    \centering
    \begin{tabular}{ccc}
        \includegraphics[width=0.3\textwidth]{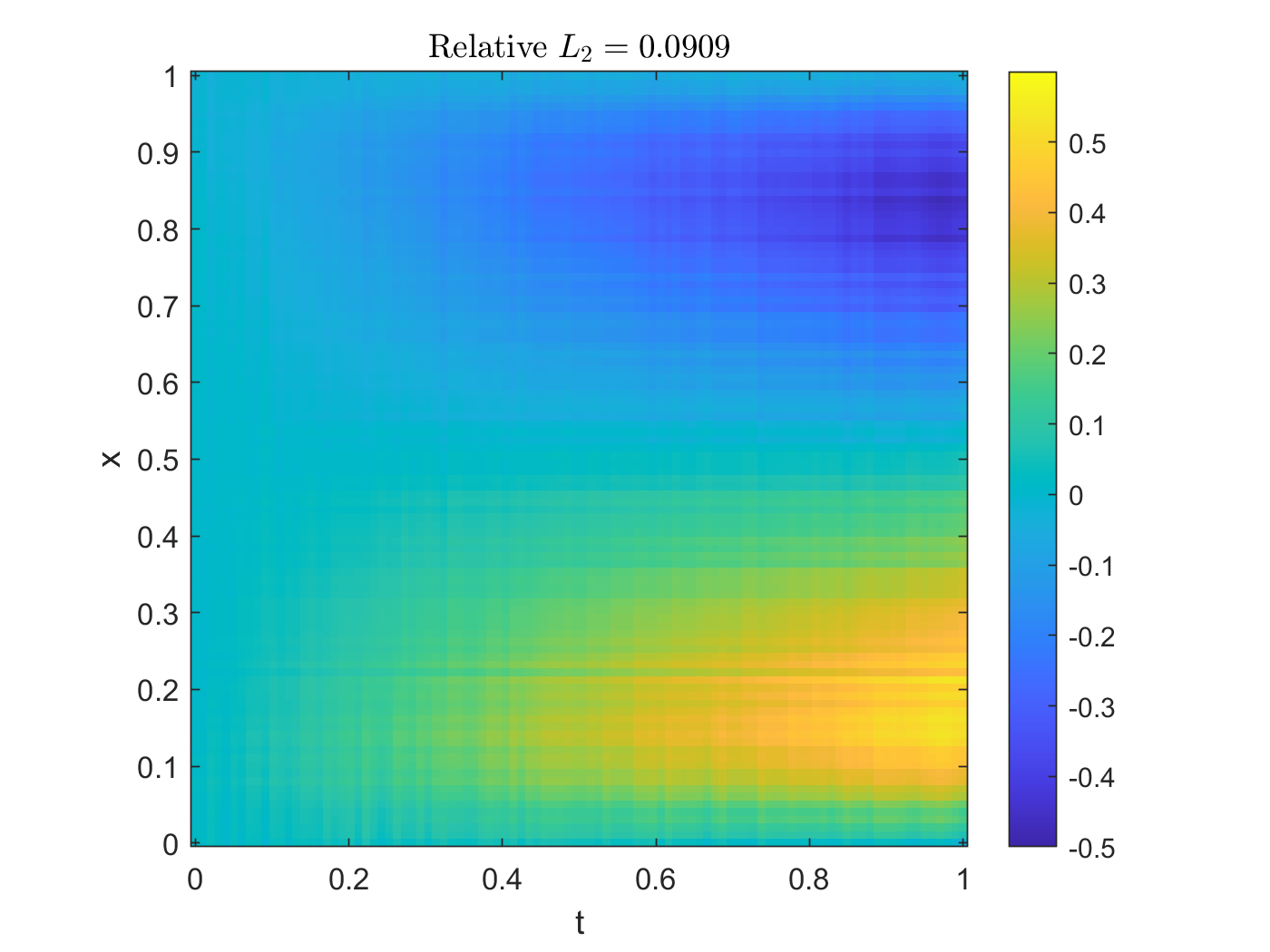} &
        \includegraphics[width=0.3\textwidth]{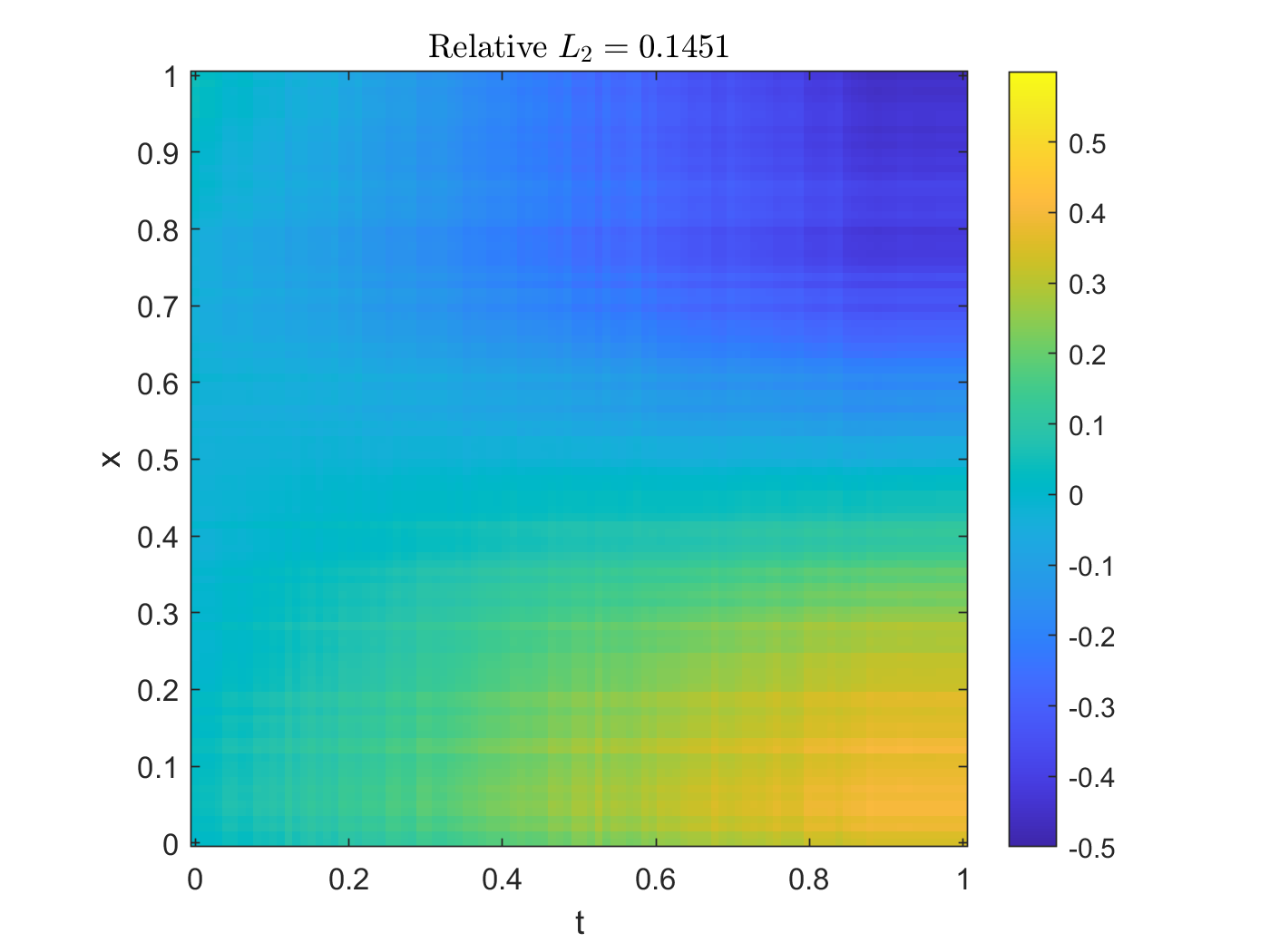} &
        \includegraphics[width=0.3\textwidth]{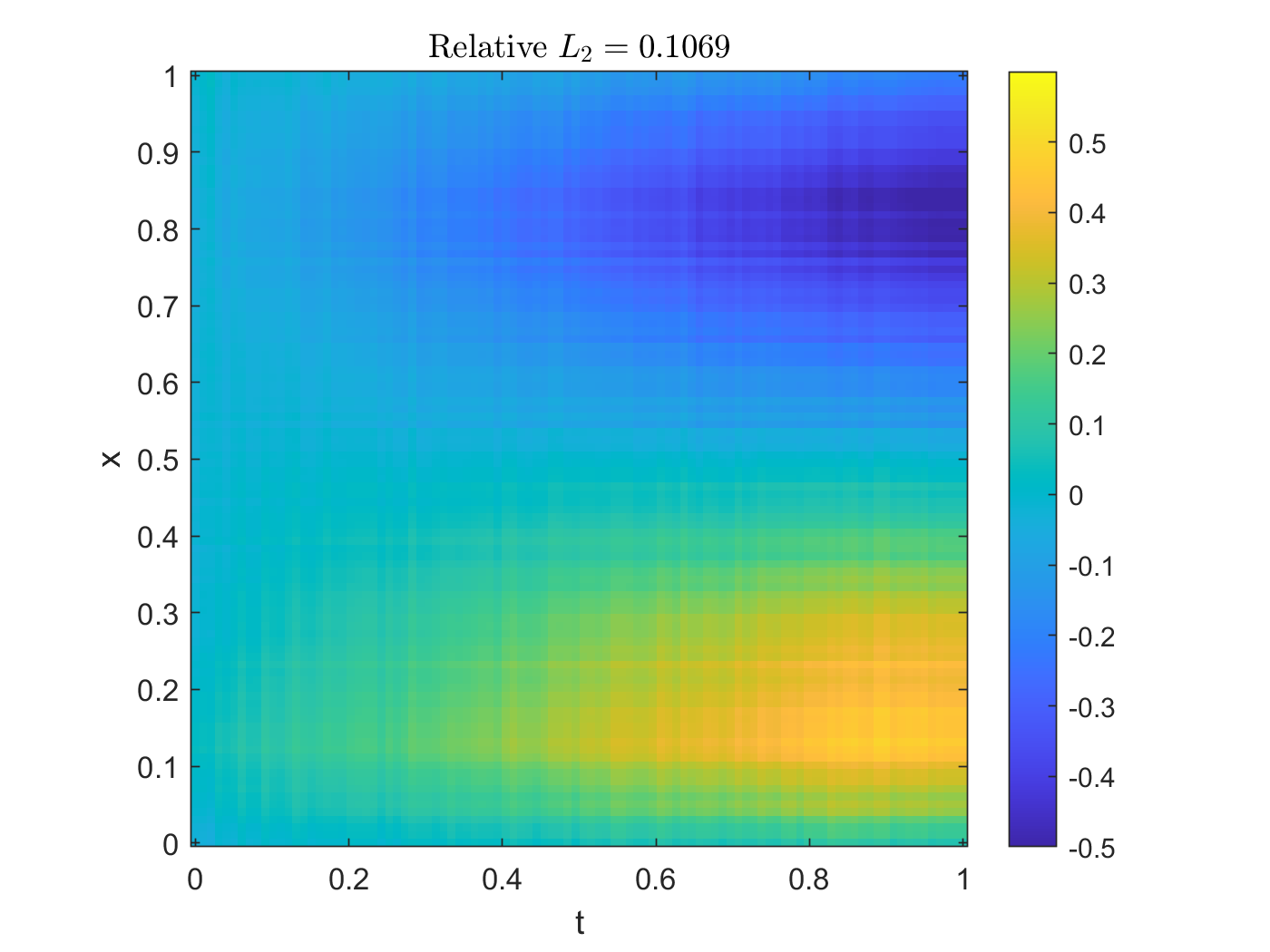} \\
        (a) Prediction(SG-BP) &  (b) Prediction(SG-RFG-G) & (c) Prediction(SG-RFG-L) \\
        \includegraphics[width=0.3\textwidth]{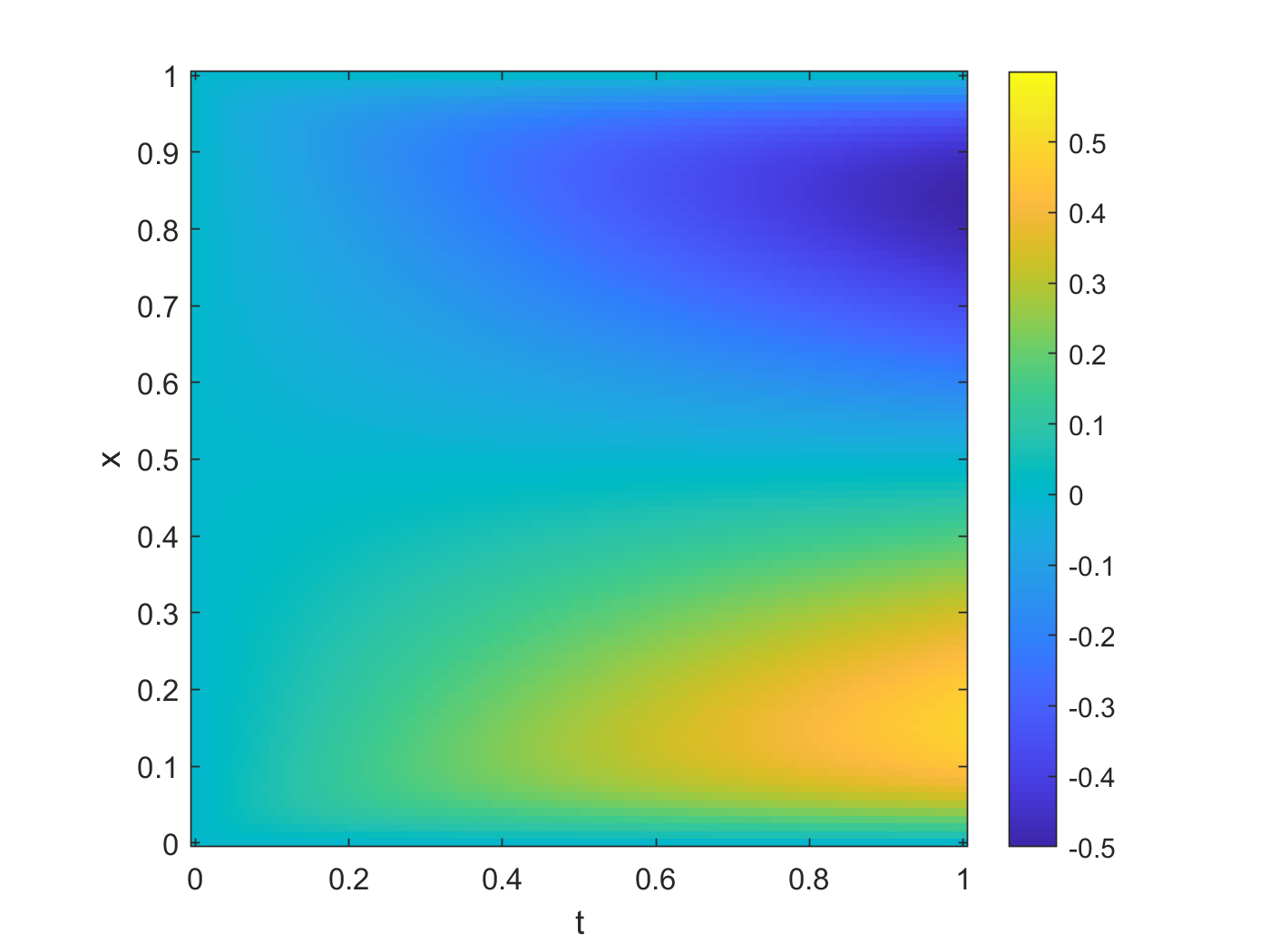} &
        \includegraphics[width=0.3\textwidth]{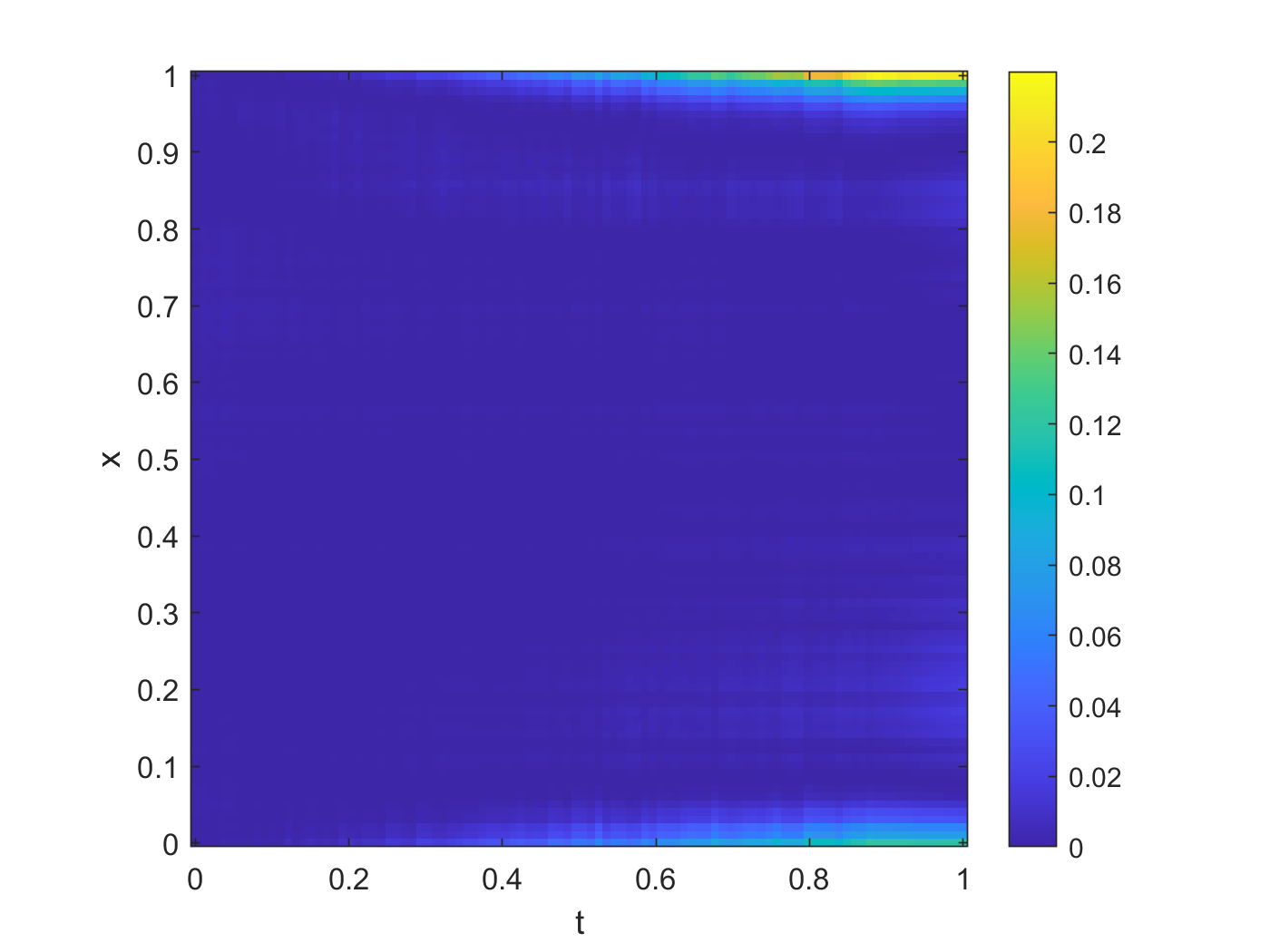} &
        \includegraphics[width=0.3\textwidth]{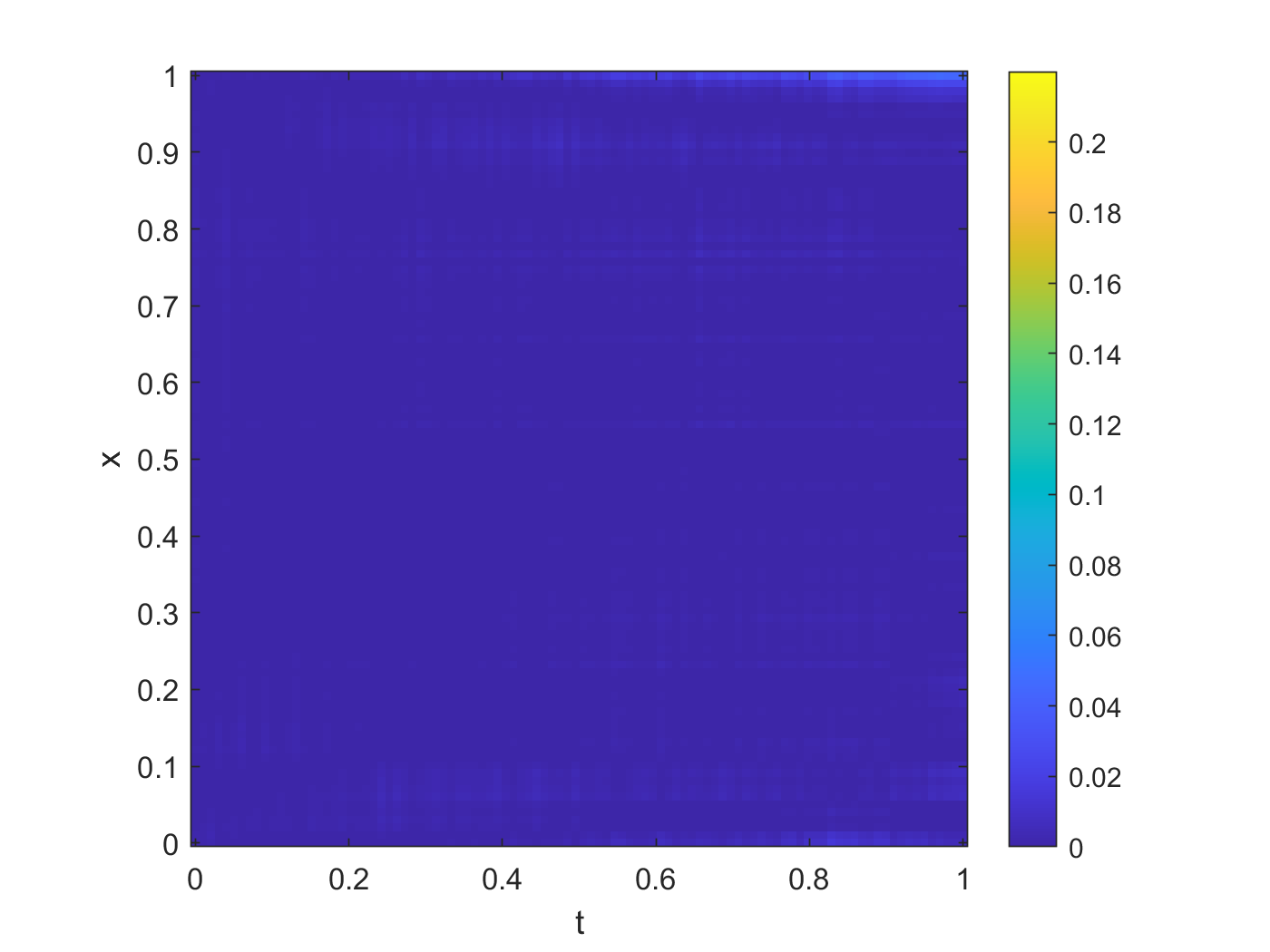} \\
        (d) Ground truth &  (e) Point-wise error(SG-RFG-G) & (f) Point-wise error(SG-RFG-L)
        
    \end{tabular}
    \caption{Solution of diffusion-reaction PDE.}
    \label{fig:dr}
\end{figure}

Finally, we solve a nonlinear reaction-diffusion PDE of the following form:
$$\frac{\partial u}{\partial t} = D \frac{\partial^2 u}{\partial x^2} + k u^2 + f(x), \quad x \in (0,1), \ t \in (0,1]$$ 
with zero initial and boundary conditions, where $D=0.01$ is the diffusion coefficient and $k=0.01$ is the reaction rate. This could also be learned using only 1 hidden layer with width 16 by the spiking version of SepONet. The simulation time step is 32. We sample 200 functions $f$ from a GRF with a Gaussian kernel with length scale 1. By solving the equation numerically, we have 200 samples and we use half for training and the other half for test. The grid is $100 \times 100$. Figure \ref{fig:dr} shows the prediction of the solution and the point-wise error of using global perturbation and layer-wise perturbation in randomized forward gradients for a sample. We see that the prediction result using layer-wise RFG achieves similar results with back-propagation.

\textbf{Computational cost.} 
During neural network training, the computation cost can be measured by counting the basic mathematical operations needed to process data. In general, traditional back-propagation \cite{gomez2017reversible} requires:

$\bullet{\text {A forward pass through all layers to compute outputs.}}$

$\bullet{\text {A backward pass that needs twice as many operations as the forward pass to calculate gradients.}}$

Our proposed randomized forward-mode gradient approach for SNNs simplifies the training by requiring only a single forward pass, eliminating the need for backward propagation entirely. This reduces the total computational cost in training by approximately 66\% compared to traditional back-propagation.
In our current implementation, we apply weight perturbations in one direction per training iteration. With specialized sampling hardware, multiple perturbation directions could be applied simultaneously, potentially offering even significant computational efficiency gains.


\section{Summary}
We presented a back-propagation-free training method for SNNs, employing randomized forward-mode gradient combined with various perturbation methods. All experiments were conducted with relatively small SNNs, achieving satisfactory results. Due to the inherent sparsity of SNNs, these computations are expected to be energy-efficient. The proposed randomized forward-mode gradient approach for SNNs achieves comparable accuracy to back-propagation but with reduced computational costs.

Future work will focus on exploring multi-directional perturbations and implementing this method on Intel's Loihi-2 hardware to further enhance energy efficiency.


{\small
\bibliographystyle{unsrt}
\bibliography{ref}
}

\end{document}